\documentclass[table, 11pt]{article}
\usepackage[table, svgnames, dvipsnames]{xcolor} % Gộp xcolor
\usepackage{tikz}
\usepackage[justification=centering]{caption}
\usepackage[most]{tcolorbox}
\colorlet{promptbg}{blue!15}
\usepackage[colorlinks=true, citecolor=blue, linkcolor=blue, urlcolor=blue]{hyperref}
 \usepackage[final]{acl}
\usepackage{amsfonts} % For \mathbb{R}
\usepackage{booktabs} % For better table rules
\usepackage{multirow} % If needed for table structure, though not strictly for this one
\usepackage{graphicx} % ở phần preamble nếu chưa có
\usepackage{amsfonts} % For \mathbb{R}
\usepackage{booktabs} % For professional tables
\usepackage{multirow} % For multi-row cells in tables
\usepackage{graphicx} % Required for includegraphics (though not used in text here)
\usepackage{xcolor} % For coloring text
\usepackage{bold-extra} % For bold text if needed, or just use \textbf
\usepackage{times}
\usepackage{latexsym}
\usepackage{enumitem}
\usepackage[T1]{fontenc}
\usepackage[utf8]{inputenc}
\usepackage{microtype}
\usepackage{inconsolata}
\usepackage{graphicx}
\usepackage{amsmath}
\usepackage{amsfonts}
\usepackage{bm}
\usepackage{multirow}
\usepackage{booktabs}
\usepackage{xspace}
\usepackage{adjustbox}
\usepackage{float}
\usepackage{stfloats}
\usepackage{natbib} % Thêm natbib để hỗ trợ \citep, \citet

\usepackage{colortbl}
\usepackage{subfigure}

\usepackage[linesnumbered,ruled,lined]{algorithm2e}

\definecolor{lightgray}{gray}{0.80}
\definecolor{darkgreen}{rgb}{0.0, 0.5, 0.0}
\newcolumntype{g}{>{\columncolor{lightgray}}c}

\title{MTA: Multi-Granular Trajectory Alignment for Large Language Model Distillation}

\author{
    \textbf{Pham Khanh Chi\textsuperscript{1}\footnotemark[1]},
  \textbf{Quoc Phong Dao\textsuperscript{1}\footnotemark[1]},   
   \textbf{Thuat Nguyen\textsuperscript{1}}, \\
    \textbf{Linh Ngo Van\textsuperscript{1,\dag}},
    \textbf{Trung Le\textsuperscript{2}},
  \textbf{Thanh Nguyen\textsuperscript{3}}  
  \bigskip \\
\textsuperscript{1}Hanoi University of Science and Technology, \\
\textsuperscript{2}Monash University,
\textsuperscript{3}University of Oregon
}

\begin{document}

\maketitle
\renewcommand{\thefootnote}{\fnsymbol{footnote}}
\footnotetext[1]{Equal contribution}
\footnotetext[2]{Corresponding author: \href{mailto:email@domain}{ linhnv@soict.hust.edu.vn}}
\renewcommand*{\thefootnote}{\arabic{footnote}}
\begin{abstract}
    % Cross-Tokenizer Knowledge Distillation (CTKD) enables knowledge transfer between a large language model and a smaller student, even when they employ different tokenizers. While existing approaches mainly focus on token-level alignment strategies, which are often brittle and sensitive to discrepancies between tokenizers, we argue that the method of aggregating tokens into more robust representations before distillation is of equal importance. In this paper, we introduce \textbf{SRA} (\textbf{S}pan \textbf{R}epresentation \textbf{A}lignment for Large Language Model Distillation), a novel framework that reframes CTKD through the physical lens of Multi-Particle Dynamical Systems. SRA shifts the fundamental unit of alignment from tokens to robust, tokenizer-agnostic spans. We model each span as a cluster of particles and represent its state by its Center of Mass (CoM) - an attention-weighted average that captures rich semantic information. We leverage the concept of span centers of mass with attention-derived weighting to prioritize the most salient spans. In addition, we employ a geometric regularizer to preserve the structural integrity of the representation space and introduce aligned span logit distillation to enhance knowledge transfer across models. In challenging cross-architecture distillation experiments, SRA consistently and significantly outperforms state-of-the-art CTKD baselines, validating our physically-grounded approach.

Knowledge distillation is a key technique for compressing large language models (LLMs), but most existing methods align representations at fixed layers or token-level outputs, ignoring how representations evolve across depth. As a result, the student is only weakly guided to capture the teacher’s internal relational structure during distillation, which limits knowledge transfer. To address this limitation, we propose \textbf{Multi-Granular Trajectory Alignment (MTA)}, a framework that aligns teacher and student representations along their layer-wise transformation trajectory. MTA adopts a layer-adaptive strategy: lower layers are aligned at the word level to preserve lexical information, while higher layers operate on phrase-level spans (e.g., noun and verb phrases) to capture compositional semantics. We instantiate this idea through a Dynamic Structural Alignment loss that matches the relative geometry among semantic units within each layer. This design is motivated by empirical findings that Transformer representations become increasingly abstract with depth, and is also consistent with linguistic views in which higher-level meaning emerges through the composition of lower-level lexical units. We further incorporate a Hidden Representation Alignment loss to directly align selected teacher–student layers. Experiments show that MTA consistently outperforms state-of-the-art baselines on standard benchmarks, with ablations confirming the contribution of each component.

\end{abstract}
\section{Introduction}\label{sec:intro}
While Large Language Models (LLMs) excel in NLP tasks, their computational cost necessitates compression techniques \citep{xu2024surveyknowledgedistillationlarge}. To address this problem, Knowledge Distillation (KD) \citep{hinton2015distillingknowledgeneuralnetwork} has emerged as a pivotal model compression paradigm, where a smaller "student" model is trained to mimic the behavior of a larger, more capable "teacher" model. Conventional KD approaches for LLMs often focus on minimizing the divergence between the output probability distributions of the teacher and the student \cite{agarwal2024onpolicy, ko2024distillm, le2025token, anshumann2025sparse, truong2026ctpd}. While effective, these methods overlook the structural knowledge embedded within the intermediate representations. To address this, feature-based distillation methods have been proposed to align internal hidden states or attention maps of the student with those of the teacher \citep{jiao2020tinybert, wang2020minilmdeepselfattentiondistillation, gong2025beyond, vu2026dwa}. However, most existing approaches adopt a static and uniform alignment strategy across layers, typically operating at the token level. By treating tokens largely in isolation and applying the same objective uniformly, these methods fail to account for differences in token importance and semantic roles across layers, which can harm generalization.

Crucially, this uniform, token-level treatment is misaligned with the hierarchical nature of human language. Linguistic research has long established that natural language is organized hierarchically, composing discrete lexical units into nested phrases and higher-level semantic structures \cite{chomsky1965aspects, crain1987structure, hale2018finding}. Prior analyses of BERT have identified a functional hierarchy in which lower layers emphasize surface-level and lexical features, while higher layers encode increasingly abstract semantic information \citep{tenney2019bert, rogers2020primer, clark2019what}. Recent studies on modern LLMs have refined this distinction, characterizing lower layers as memory units for lexical and factual retrieval \citep{wang2025think, yang2025internal}, while higher layers perform abstract reasoning and manage complex subtasks \citep{yang2025internal}. As a consequence, while focusing on token-level alignment may allow the student to reproduce teacher activations at isolated layers, it fails to capture the progressive process through which representations are composed from lexical units into more abstract semantic structures, limiting generalization in downstream tasks. 

To bridge this gap, we propose \textbf{Multi-Granular Trajectory Alignment (MTA)}, an extensible framework designed to align the evolutionary trajectory of representations between teacher and student models. Our approach is motivated by linguistic principles and supported by established findings in Transformer interpretability. These observations suggest that representations in LLMs evolve in a hierarchical, bottom-up manner: lower layers process fine-grained lexical dependencies (analogous to leaf nodes), whereas higher layers synthesize these inputs into abstract semantic structures (analogous to internal phrase nodes). Guided by this perspective, MTA adopts a layer-adaptive alignment strategy.
Specifically, we introduce a Dynamic Structural Alignment ($\mathcal{L}_{DSA}$) objective that aligns fine-grained word-level spans at lower layers to ground lexical foundations, and coarser phrase-level spans (e.g., Noun Phrases, Verb Phrases) at higher layers to capture compositional semantics. By explicitly modeling these multi-granular relationships, MTA compels the student to mimic not just the teacher's final state, but the geometric dynamics of its information processing. Furthermore, we incorporate a Hidden Representation Alignment ($\mathcal{L}_{Hid}$) loss to ensure precise transfer via a weighted projection mechanism.

Our main contributions are summarized as follows:
\begin{itemize}
    \item We identify the limitations of uniform intermediate alignment and propose \textbf{MTA}, a novel framework that aligns the representational trajectory of LLMs by leveraging their inherent hierarchical structure.
    \item We introduce a layer-adaptive distillation objective that transitions from word-level alignment in lower layers to phrase-level structural alignment in higher layers, reflecting the bottom-up compositional properties of language.
    \item We demonstrate that MTA serves as a generalized module that can be integrated into state-of-the-art distillation methods that share tokenization. Extensive experiments on standard instruction-following benchmarks show that MTA consistently boosts performance across diverse model architectures .
\end{itemize}
\section{Background}

This section introduces the background concepts underlying our approach.
We first review the fundamentals of knowledge distillation for autoregressive language models,
then discuss parse trees and feature dynamics distillation as perspectives that motivate our depth-wise alignment.
An extended discussion of related work is deferred to Appendix~\ref{sec:related_work}.

\subsection{Knowledge Distillation Fundamentals}

Knowledge Distillation (KD) \cite{hinton2015distillingknowledgeneuralnetwork} transfers knowledge from a teacher to a student by minimizing the divergence between their conditional distributions. For autoregressive LMs, the sequence-distribution KL can be written as a sum over token-level conditionals; in practice it can be optimized under teacher forcing on a dataset (or sampled sequences), yielding a tractable token-wise KL objective. Specifically, for a dataset $\mathcal{D}$, the loss is aggregated at each time step $t$:

\begin{equation}
    D_{\text{KL}}(p, q_\theta)  = \mathbb{E}_{\mathbf{x}} \mathbb{E}_{\mathbf{y} \sim p(\cdot|\mathbf{x})} \left[ \log \frac{p(\mathbf{y}|\mathbf{x})}{q_\theta(\mathbf{y}|\mathbf{x})} \right]
\end{equation}
{\small
\begin{equation}
     \approx \frac{1}{|\mathcal{D}|} \sum_{(\mathbf{x}, \mathbf{y}) \in \mathcal{D}} \sum_{t=1}^{|\mathbf{y}|} \sum_{y_t \in V} p(y_t | \mathbf{y}_{<t}, \mathbf{x}) \log \frac{p(y_t | \mathbf{y}_{<t}, \mathbf{x})}{q_\theta(y_t | \mathbf{y}_{<t}, \mathbf{x})}
    \label{eq:token_kld}
\end{equation}} where $V$ is the vocabulary and $\mathbf{y}_{<t}$ denotes the token history. This formulation enables the student to approximate the teacher's distribution and capture the inherent ``dark knowledge'' in inter-class correlations \cite{kim2016sequencelevelknowledgedistillation, agarwal2024onpolicy}. 

\subsection{Parse Tree}
A parse tree structurally represents the hierarchical nature of natural language, organizing flat text into meaningful syntactic units to capture semantic compositionality \cite{socher2011parsing, socher2013recursive}. In recursive deep learning, this topology guides the bottom-up merging of representations from leaf nodes to the root \cite{tai2015improved}. Motivated by the hierarchical structure of language captured by parse trees, we treat the depth-wise evolution of Transformer representations as loosely analogous to bottom-up compositional construction, and design a layer-adaptive alignment strategy accordingly as shown in Figure~\ref{fig:parse_tree}. 
\subsection{Feature Dynamics Distillation}
\label{sec:background_fdd}
% {\color{red} Thanh: Do we need this section 2.3 for the Background?  Perhaps, it is best to move this section 2.3 together with DistiLLM losses in appendix.}
%viết thêm motivation để cho cái này vào 
Feature Dynamics Distillation (FDD) \citep{gong2025beyond} extends KD by viewing Transformer depth as discrete time steps of a continuous-depth dynamical system, following ODE-inspired interpretations of deep networks \citep{chen2018neuralode, lu2019understanding}. Under this view, distillation should match not only intermediate states but also their layer-wise evolution. To handle dimension mismatch, FDD maps intermediate hidden states $h_l$ through each model's LM head into the vocabulary space, denoted as $y_l = \log f_{\text{head}}(h_l)$. It then optimizes (i) a \textbf{Trajectory Loss} that aligns intermediate predictive distributions at selected layers, and (ii) a \textbf{Derivative Loss} that aligns finite-difference updates $\Delta y_l = y_l - y_{l-1}$ using cosine distance:
\begin{align}
\mathcal{L}_{\text{Traj}} &= \sum_{j} \mathcal{D}_{KL}\!\left(P(y^{\mathcal{T}}_{l_j}) \,\|\, P(y^{\mathcal{S}}_{l_j})\right),\\
\mathcal{L}_{\text{Der}}  &= \sum_{j} \left(1 - \text{Cos}(\Delta y^{\mathcal{T}}_{l_j}, \Delta y^{\mathcal{S}}_{l_j})\right).
\end{align}
While effective for capturing continuity in prediction dynamics, FDD aligns prediction dynamics using the same LM-head-based objective at the selected layers, without explicitly encoding depth-dependent linguistic structure or salience-aware token weighting \citep{gong2025beyond}. Inspired by the trajectory-based distillation framework of \citet{gong2025beyond}, our approach departs from it by explicitly aligning the \emph{evolution of relational geometry} across network depth. 
Rather than matching prediction trajectories alone, we leverage hierarchical structure to guide depth-aware alignment of token relations, enabling more semantically faithful knowledge transfer.
\section{Methodology}
\label{sec:proposed_method}

In this section, we introduce \textbf{Multi-Granular Trajectory Alignment (MTA)}, a modular framework designed to enhance the distillation of Large Language Model (LLM). Unlike conventional methods that rely on uniform layer-to-layer mapping, MTA adopts a depth-aware alignment strategy that operates at different semantic granularities across network depth. MTA proposes two complementary objectives: Dynamic Structural Alignment ($\mathcal{L}_{\text{DSA}}$), which preserves the relational geometry among semantic units (words and phrases) within each layer, and Hidden Representation Alignment ($\mathcal{L}_{\text{Hid}}$), which enforces feature-level consistency between teacher and student representations. Together, these objectives constrain both the internal structure of representations within each layer and their transformation across depth.

\subsection{Motivation: The Hierarchical Representational Trajectory}
\label{subsec:motivation}

\begin{figure}[t]
    \centering
    \includegraphics[width=\linewidth]{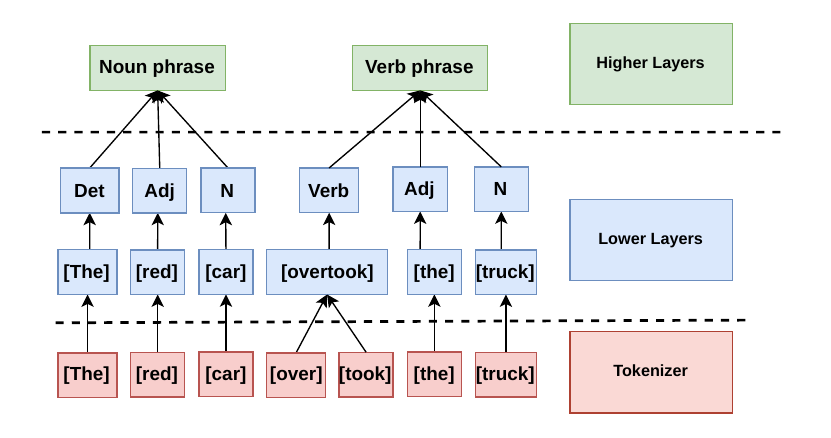}
    \caption{The correspondence between linguistic compositionality and the layer-wise evolution of representations in large language models.}
    \label{fig:parse_tree}
\end{figure}

Most existing Knowledge Distillation (KD) methods match representations at fixed, static points, such as final logits or intermediate hidden states and attentions
\citep{hinton2015distillingknowledgeneuralnetwork, sun2019patientknowledgedistillationbert, jiao2020tinybert, truong2025emo, hoang2026mcw, xu2024surveyknowledgedistillationlarge}.
While effective for transferring local signals, these approaches largely ignore how representations evolve across network depth.

Interpretability studies consistently show that Transformer layers exhibit a hierarchical organization.
Lower layers emphasize surface-level and lexical information, whereas higher layers progressively encode abstract semantic and compositional properties
\citep{tenney2019bert, rogers2020primer, clark2019what}.
Recent analyses of LLMs further characterize lower layers as repositories for factual and lexical memory \citep{wang2025think}, and higher layers as supporting abstract reasoning and compositional inference \citep{yang2025internal}.

We therefore view a model’s internal representations as forming a \textbf{hierarchical representational trajectory}: an ordered sequence of representation spaces whose semantic granularity systematically changes with depth.
Under this perspective, applying a single, uniform alignment rule across all layers is suboptimal.

 To respect this hierarchical trajectory during distillation, MTA uses \textit{word-level spans} for alignment in lower layers to preserve lexical grounding, and \textit{phrase-level spans} (e.g., noun and verb phrases) in higher layers to capture compositional structure. This layer-adaptive design aims to better reflect the depth-wise transformation of representations than a single, uniform matching rule. Figure~\ref{fig:parse_tree} illustrates the connection between linguistic compositionality and our depth-adaptive alignment strategy.

% To respect this hierarchical trajectory during distillation, MTA uses \textit{word-level spans} for alignment in lower layers to preserve lexical grounding, and \textit{phrase-level spans} (e.g., noun and verb phrases) in higher layers to capture compositional structure. This layer-adaptive design aims to better reflect the depth-wise transformation of representations than a single, uniform matching rule.
% To respect this hierarchical trajectory during distillation, MTA adapts the granularity of alignment according to layer depth:
% \begin{itemize}
%     \item \textbf{Lower layers.} Representations are aligned using \textit{word-level spans}, encouraging the student to preserve fine-grained lexical relationships and syntactic precision.
%     \item \textbf{Higher layers.} Alignment is performed over \textit{phrase-level spans} (e.g., noun and verb phrases), relaxing strict word-to-word matching in favor of capturing broader relational structure among abstract semantic units.
% \end{itemize}

% Conceptually, our approach differs from prior trajectory-based distillation methods that align representational derivatives across layers \citep{gong2025beyond}.
% Instead, we align the \emph{evolution of relational geometry} along network depth, providing a depth-aware and semantically adaptive supervision signal.
% Figure~\ref{fig:parse_tree} illustrates the correspondence between linguistic compositionality and our adaptive layer-wise alignment strategy.
\subsection{Layer-Adaptive Multi-Granular Spans}
\begin{figure*}
    \centering
    \includegraphics[width=1.0\linewidth]{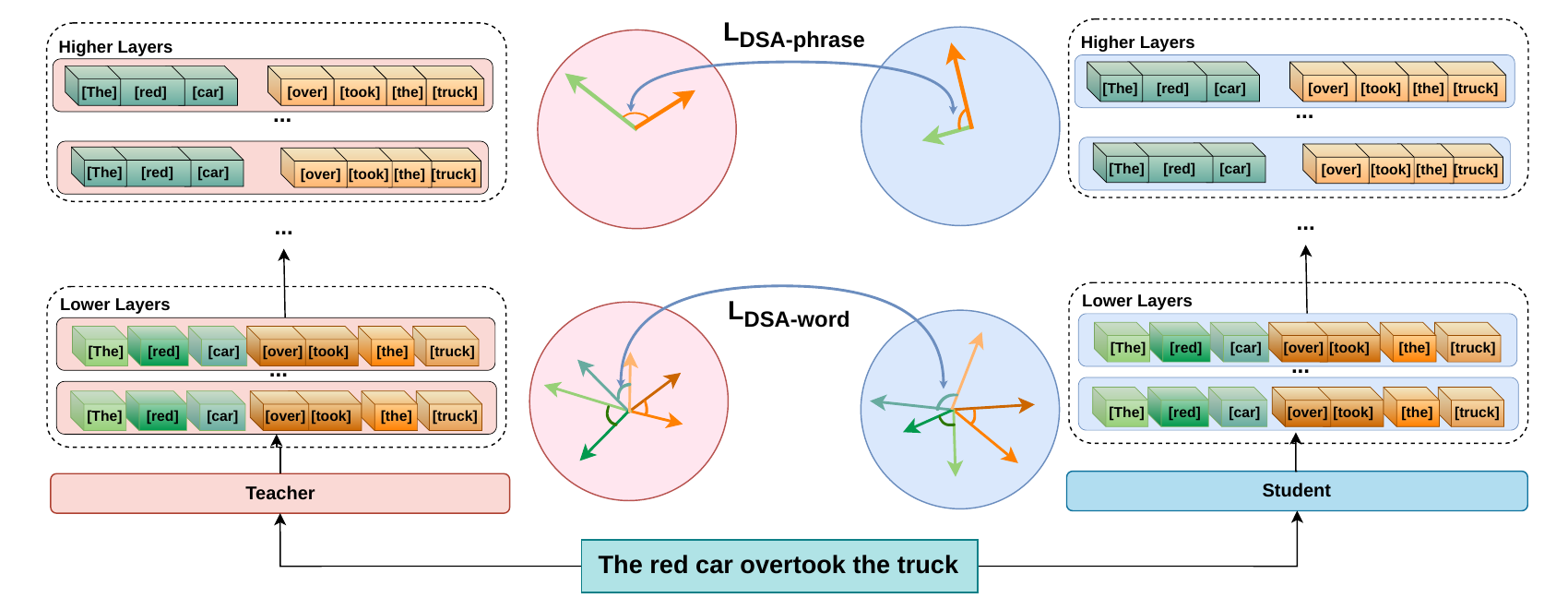}
    \caption{\textbf{Dynamic Structural Alignment ($\mathcal{L}_{\text{DSA}}$).} This objective enforces geometric consistency. It calculates the pairwise relational distances between semantic spans (words or phrases) within a layer for both Teacher and Student. By minimizing the discrepancy between these two topological structures across network depths, the Student learns to replicate the Teacher's representational trajectory.}
    \label{fig:DSA}
\end{figure*}

To operationalize this hierarchical alignment, we define semantic units tailored to different layers. We describe how we divide the selected key layers into \textit{Lower} and \textit{Higher} functional groups, detailed in Appendix~\ref{layer_mapping}. To implement this, we utilize a syntactic parser (e.g., \texttt{spaCy}) to extract the corresponding spans from the input text. 
% While in our experiments, we employ \texttt{spaCy} for its robust performance in English, the framework accommodates any span extraction method, from statistical chunkers for low-resource languages to specialized parsers for code.

 \paragraph{Token Weight.}
% The importance of each token is estimated based on how strongly it is attended to by other tokens in the sequence. To ensure stable and comparable similarity scores, token representations at layer $l$ are first normalized by their feature-wise standard deviation. We then apply a pairwise self-attention mechanism in which each token measures its relevance to all other tokens using scaled dot-product similarity. A masking strategy is employed to exclude self-attention and padding tokens, forcing the model to rely solely on contextual information from surrounding tokens. After softmax normalization, the attention matrix reflects how much contextual focus each token receives. The final importance weight for a token is computed as the average attention it receives from all other tokens, providing an interpretable scalar that captures its overall contextual influence within the sequence. Span importance is then computed by aggregating the Teacher's token-level weights within each span, ensuring that spans containing more salient tokens receive higher influence during alignment.
Due to the unidirectional nature of causal attention in autoregressive LLMs, native attention weights are inherently biased toward earlier tokens, which tend to accumulate higher attention scores simply because they are visible to more subsequent positions when calculated based on attention centrality (or aggregated received attention). To bypass the unidirectional constraints and capture bidirectional dependencies, we compute token weights using a self-attention mechanism without self-loops.
Let $\mathbf{H}_l \in \mathbb{R}^{N \times d}$ denote the hidden states at layer $l$. We first standardize token representations:
\begin{equation}
\hat{H}_{t,l} = \frac{H_{t,l}}{\sigma(H_{t,l})}.
\end{equation} Pairwise attention scores with diagonal and padding masking are computed as:
\begin{equation}
S_{s\to t,l} = \frac{\hat{H}_{s,l} \hat{H}_{t,l}^\top}{\sqrt{d}} + M_{s,t}.
\end{equation} The attention weights $\alpha_{s \to t,l}$ are then obtained by applying a softmax over the destination token dimension of the attention score matrix $S_l$. After that we compute the importance score of token $t$:

\begin{equation}
w_{t,l} = \frac{1}{N} \sum_{s=1}^{N} \alpha_{s\to t,l}.
\end{equation}
% The importance of token $t$ is defined as the average attention it receives:
% \begin{equation}
% w_{t,l} = \frac{1}{N} \sum_{s=1}^{N} \alpha_{s\to t,l}.
% \end{equation}
These weights are used for span-level aggregation in subsequent objectives. The more detailed explanation is in Appendix \ref{sec:appendix_token_weights}.
\paragraph{Span Definitions.}
At \textbf{lower layers}, we group tokens into complete \textit{Word Spans} to capture fine-grained lexical processing. At \textbf{higher layers}, we extract \textit{Noun Phrases (NPs)} and \textit{Verb Phrases (VPs)} to serve as \textit{Phrase Spans}, which are widely used as meaningful syntactic constituents in chunking and predicate-argument semantics \citep{ramshaw1995chunking, gildea2002srl}. Spans are extracted over the entire input-output sequence, including both the prompt and the generated response. For each span $k$ at layer $l$, we compute its representation $U_{k,l}$ as a weighted average of constituent token hidden states $H_{t,l}$:
\begin{equation}
    U_{k,l} = \frac{\sum_{t \in S_k} w_{t,l} H_{t,l}}{\sum_{t \in S_k} w_{t,l}}
    \label{eq:span_rep}
\end{equation}
where $S_k$ is the set of tokens in span $k$, and $w_{t,l}$ is the importance weight of token.

\paragraph{Span Weight.}
The weight of a span is computed by aggregating the weight of its constituent tokens and normalizing across spans.
For span $i$ at layer $l$, we define:
\begin{equation}
w^{\text{sp}}_{i,l}
=
\frac{\tilde{w}^{\text{sp}}_{i,l}}
{\sum_{j=1}^{N^{\text{sp}}_{l}} \tilde{w}^{\text{sp}}_{j,l}},
\qquad
\tilde{w}^{\text{sp}}_{i,l}
=
\sum_{t=s^{l}_{i}}^{e^{l}_{i}} w^{\mathcal{T}}_{t,l}.
\label{eq:span_weight}
\end{equation}

where $[s_i^l, e_i^l]$ denotes the token indices covered by span $i$, $w^T_{t,l}$ is the Teacher-derived token importance weight, and $N^{\text{sp}}_l$ is the number of spans at layer $l$.

\subsection{Dynamic Structural Alignment Loss ($\mathcal{L}_{\text{DSA}}$)}

The central objective of MTA is the \textbf{Dynamic Structural Alignment Loss}. This loss enforces geometric consistency between the Teacher and Student span representations, as shown in Figure \ref{fig:DSA}. It operates on a set of selected key layers ${L}_{\text{key}}$ (spanning both lower/word-level and higher/phrase-level layers). We formulate $\mathcal{L}_{\text{DSA}}$ to minimize the discrepancy in pairwise distances between spans within a layer:

\begin{equation}
\label{eq:dsa_loss}
    \mathcal{L}_{\text{DSA}}= \frac{1}{\|{L}_{\text{key}}\|}\sum_{l \in{L}_{\text{key}}} \mathcal{L}_{\text{DSA}}^{(l)}
\end{equation}

{\small
\begin{equation}
\label{eq:dsa_loss_layer}
    \mathcal{L}_{\text{DSA}}^{(l)} = \sum_{i=1}^{N^{\text{sp}}_l} \sum_{j=i+1}^{N^{\text{sp}}_l} w_{ij,\phi(l)}^{\text{sp}} \Big( d(U_{i,l}^{\mathcal{S}}, U_{j,l}^{\mathcal{S}}) - d(U_{i,\phi(l)}^{\mathcal{T}}, U_{j,\phi(l)}^{\mathcal{T}}) \Big)^2
\end{equation}
}

where $d(\cdot,\cdot)$ is cosine distance, $N^{\text{sp}}_l$ is the number of spans at layer $l$ and $\phi(\cdot)$ denote layer mapping function. We weight span pairs by salience, $w_{ij,\phi(l)}^{\text{sp}} = w_{i,\phi(l)}^{\text{sp}} w_{j,\phi(l)}^{\text{sp}}$, where $w_{k,\phi(l)}^{\text{sp}}$ aggregates token weights within span $k$. This prioritizes alignment between semantically salient spans. Minimizing $\mathcal{L}_{\text{DSA}}$ encourages the student to match the teacher's within-layer relational structure and its evolution across depth.
% note về teacher attention weight chứ không phải cả teacher cả student
\subsection{Hidden Representation Alignment ($\mathcal{L}_{\text{Hid}}$)}

% có các paper relation kd về việc nên distil
\begin{figure}
    \centering
    \includegraphics[width=1.0\linewidth]{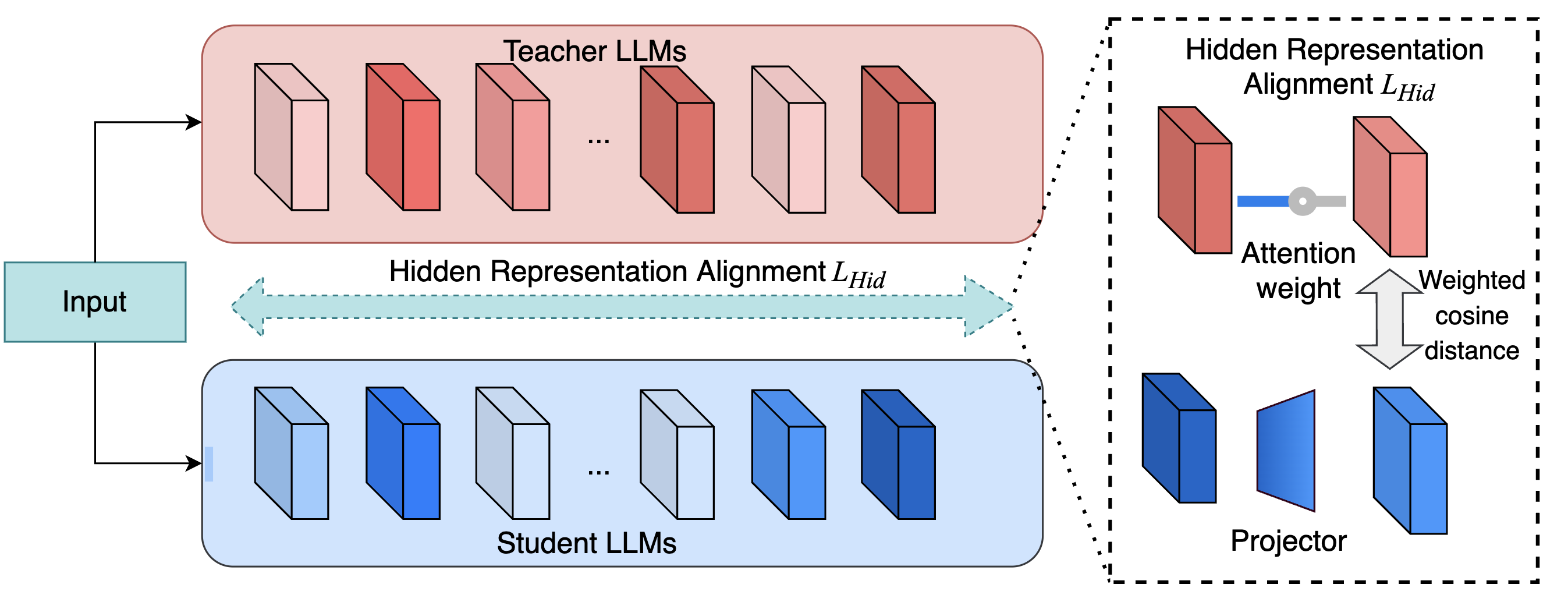}
    \caption{\textbf{Hidden Representation Alignment Strategy.} The student learns to match teacher representations using a weighted cosine distance objective, ensuring accurate feature reconstruction at key layers.}
    \label{fig:hidden_rep}
\end{figure}
% {\color{red} Thanh: Can we use this loss function at span levels as well?} 
To complement the structural constraints, we incorporate a feature-level loss to align specific hidden states. Since the Student dimension ($d_\mathcal{S}$) is typically smaller than the Teacher dimension ($d_\mathcal{T}$), we employ a learnable linear projector $W_l \in \mathbb{R}^{d_\mathcal{S} \times d_\mathcal{T}}$ to map the Student's representations into the Teacher's space:
\begin{equation}
    \tilde{H}^{\mathcal{S}}_{t,l} = H^{\mathcal{S}}_{t,l} W_l
\end{equation}
We then minimize the weighted cosine distance between the projected Student states and the Teacher states:
\begin{equation}
\mathcal{L}_{\text{Hid}}
=
\sum_{l \in \mathcal{L}_{\text{key}}}
\sum_{t \in \mathcal{M}_l}
w^{\mathcal{T}}_{t,l}
\left(
1 -
\frac{
\langle \tilde{H}^{\mathcal{S}}_{t,l}, H^{\mathcal{T}}_{t,l} \rangle
}{
\left\|
\tilde{H}^{\mathcal{S}}_{t,l}
\right\|_2
\,
\left\|
H^{\mathcal{T}}_{t,l}
\right\|_2
}
\right)
\end{equation}

where ${L}_{\text{key}}$ denotes the set of selected intermediate layers used for distillation,  $\mathcal{M}_l$ contains tokens covered by the extracted spans at layer $l$, and $w_{t,l}^\mathcal{T}$  is the Teacher-derived token importance weight defined previously. This ensures that the Student focuses on replicating the features of the most informative tokens. Figure~\ref{fig:hidden_rep} provides an overview of how the proposed loss works.

\subsection{Optimization Objective}
MTA is designed as a modular plug-in for existing distillation frameworks (e.g., DistiLLM \citep{ko2024distillm}, DistiLLM-2 \citep{ko2025distillm2}, FDD \citep{gong2025beyond}). The total training objective combines the base distillation loss $\mathcal{L}_{\text{Base}}$ of the original methods with our proposed structural and hidden losses:

\begin{equation}
    \mathcal{L}_{\text{Total}} = \mathcal{L}_{\text{Base}} + \lambda_{\text{DSA}} \mathcal{L}_{\text{DSA}} + \lambda_{\text{Hid}} \mathcal{L}_{\text{Hid}}
\end{equation}
where $\lambda_{\text{DSA}}$ and $\lambda_{\text{Hid}}$ are hyperparameters balancing structural preservation and feature alignment.
% {\color{red}Thanh: can we also enhance $\mathcal{L}_{base}$ itself with span-level representations?}

\section{Experiments}

% {\color{red}Thanh: With additional loss components, how is the runtime performance? }
\subsection{Experimental Setup}
\paragraph{Datasets.} Our evaluation spans multiple instruction-following datasets. We adopt the preprocessing procedure of \citep{zhang2024dualspaceknowledgedistillationlarge}. Distillation is trained on \textsc{Databricks-Dolly-15k}. For evaluation, we report ROUGE-L on the Dolly test split and on four out-of-distribution benchmarks - \textsc{S-NI} \citep{wang2022benchmarking}, \textsc{VicunaEval} \citep{chiang2023vicuna}, and \textsc{SelfInst} \citep{wang2023selfinstructaligninglanguagemodels}.

\paragraph{Training and Evaluation Settings.} We evaluate our method on three student models: GPT-2-120M \citep{radford2019language}, Qwen1.5-0.5B \citep{bai2023qwen}, and OPT-1.3B \citep{zhang2022optopenpretrainedtransformer}.
For each student model, we use a correspondingly larger teacher model of the same family: GPT-2-1.5B, Qwen1.5-1.8B, and OPT-6.7B. To assess the quality of the generated outputs, we employ the ROUGE-L score \citep{lin2004rouge} as our primary evaluation metric. Additional training and evaluation setup details are provided in Appendix~\ref{sec: appendix_exp}.

\paragraph{Baselines.}
We benchmark our method against several prominent frameworks that employ the same tokenizer, including FDD \citep{gong2025beyond}, DistiLLM \citep{ko2024distillm}, and DistiLLM-2 \citep{ko2025distillm2}. A detailed description of FDD is provided in Section~\ref{sec:background_fdd}, while overviews of DistiLLM and DistiLLM-2 are presented in Appendix~\ref{sec: appendix_baseline}. We integrate our proposed method into these baselines to evaluate its effectiveness as a complementary optimization strategy.

\subsection{Main Results}

\begin{table}[h!]
\centering
\footnotesize
\renewcommand{\arraystretch}{1.15}
\setlength{\tabcolsep}{4.5pt}

\begin{tabular}{l|cccccc}
\toprule
\textbf{Methods} & \textbf{Dolly} & \textbf{SelfInst} & \textbf{Vicuna} & \textbf{S-NI} & \textbf{Avg.}\\
\specialrule{1.0pt}{1.0pt}{1.0pt}

% ================= GPT-2 1.5B -> GPT-2 120M =================
\multicolumn{6}{c}{\textit{GPT-2 1.5B} $\rightarrow$ \textit{GPT-2 120M}}\\
\midrule
Teacher        & 28.71 & 15.68 & 17.00 & 28.80 & 22.55 \\
SFT            & 23.33 & 10.56 & 15.12 & 17.08 & 16.52 \\
\midrule

FDD            & 25.47 & 12.45 & 16.44 & 23.54 & 19.48 \\
\quad \textit{w/} MTA     & \underline{25.64} & \underline{13.6} & \underline{17.00} & \underline{25.75} & \underline{20.50} \\
\midrule

DistiLLM       & 25.65 & 13.39 & 16.50 & 25.28 & 20.21 \\
\quad \textit{w/} MTA   & \underline{25.77} & \underline{14.19} & 
                        \underline{16.67} & \underline{29.18} & 
                        \underline{21.45} \\
\midrule

DistiLLM\text{-}2      & 22.44 & 12.52 & 12.30 & 27.10 & 18.59 \\
\quad \textit{w/} MTA  & \underline{24.76} & \underline{14.16} & 
                        \underline{14.28} & \underline{26.54} & 
                        \underline{19.94} \\
\specialrule{1.0pt}{1.0pt}{1.0pt}

% ================= Qwen1.5 1.8B -> Qwen1.5 0.5B =================
\multicolumn{6}{c}{\textit{Qwen1.5 1.8B} $\rightarrow$ \textit{Qwen1.5 0.5B}}\\
\midrule
Teacher        & 28.23 & 19.58 & 19.59 & 34.36 & 25.44 \\
SFT            & 24.83 & 13.31 & 16.97 & 22.07 & 19.30 \\
\midrule

FDD            & 25.08 & 12.24 & 16.08 & 23.69 & 19.27 \\
\quad \textit{w/} MTA      & \underline{25.35} & \underline{13.85} & \underline{17.21} & \underline{27.25} & \underline{20.92} \\
\midrule

DistiLLM             & 25.16 & 12.90 & 15.86 & 25.28 & 19.80 \\
\quad \textit{w/} MTA  & \underline{25.61} & \underline{13.08} & \underline{16.04} & \underline{29.32} & \underline{21.01} \\
\midrule

DistiLLM\text{-}2      & 27.48 & 17.95 & 17.14 & 30.99 & 23.39 \\
\quad \textit{w/} MTA  & \underline{27.93} & \underline{18.87} & \underline{18.63} & \underline{33.49} & \underline{24.73} \\
\specialrule{1.0pt}{1.0pt}{1.0pt}

% ================= OPT 6.7B -> OPT 1.3B =================
\multicolumn{6}{c}{\textit{OPT 6.7B} $\rightarrow$ \textit{OPT 1.3B}}\\
\midrule
Teacher        & 27.60 & 16.40 & 17.80 & 30.30 & 23.03 \\
SFT            & 26.00 & 11.40 & 15.60 & 23.10 & 19.03 \\
\midrule

FDD            & 26.07 & 14.82 & 17.09 & 28.98 & 21.74 \\
\quad \textit{w/} MTA      & \underline{26.49} & \underline{16.47} & \underline{17.51} & \underline{31.12} & \underline{22.90} \\
\midrule

DistiLLM       & \underline{27.87} & 15.99 & 18.02 & 30.02 & 22.98 \\
\quad \textit{w/} MTA & 27.43 & \underline{17.07} & \underline{18.40} & \underline{32.97} & \underline{23.97} \\
\midrule

DistiLLM\text{-}2     & 26.31 & 17.62 & \underline{17.60} & 30.85 & 22.96 \\
\quad \textit{w/} MTA & \underline{26.79} & \underline{18.16} & 16.55 & \underline{31.39} & \underline{23.22} \\
\bottomrule
\end{tabular}
\caption{Performance of different distillation methods across datasets. “w/ MTA’’ denotes incorporating our proposed Multi-Granular Trajectory Alignment (MTA) strategy on top of each baseline.}
\label{tab:main_results}
\end{table}

Table~\ref{tab:main_results} provides an overview of the ROUGE-L evaluation results for all teacher-student configurations examined in this study that range from lightweight to large-scale architectures. As observed, the integration of our MTA module consistently improves the performance of all baseline frameworks, effectively narrowing the performance gap between student and teacher models. This improvement is robust across varying scales and architectures, including GPT-2, Qwen1.5, and OPT. Such scalability suggests that our approach functions as a generalized plug-and-play optimization layer.

\begin{figure}[t]
\centering
\includegraphics[width=\linewidth]{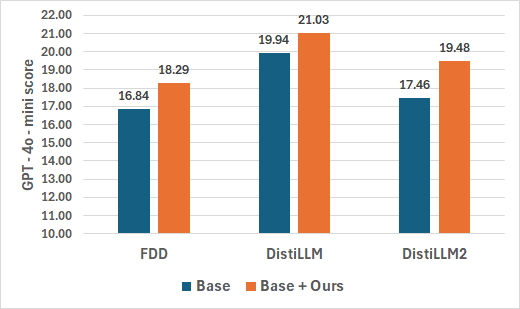}
\captionof{figure}{GPT-4o-mini evaluation scores (1-100) for distilling GPT-2 1.5B into GPT-2 120M.}
\label{fig:ChatGPT_eval_gpt2}
\end{figure}

\begin{figure}[t]
\centering
\includegraphics[width=\linewidth]{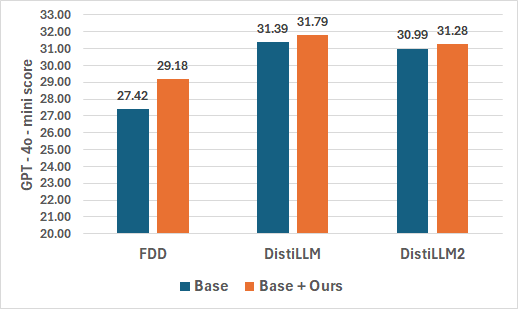}
\captionof{figure}{GPT-4o-mini evaluation scores (1-100) for distilling OPT 6.7B into OPT 1.3B.}
\label{fig:ChatGPT_eval_opt2}
\end{figure}

Beyond ROUGE-L–based metrics, we employ GPT-4o-mini as an evaluator (LLM-as-a-judge) using the prompts detailed in Appendix~\ref{appendix:prompt} to assess the semantic quality of the generated outputs. The model is prompted to score responses based on factual accuracy, coherence, and helpfulness. As illustrated in Figure~\ref{fig:ChatGPT_eval_gpt2} and Figure~\ref{fig:ChatGPT_eval_opt2}, our method yields consistent improvements in evaluation scores across both architectures. Notably, in the FDD framework, the integration of MTA leads to significant gains. Even for stronger baselines like DistiLLM and DistiLLM-2, our approach maintains a positive trajectory. These results confirm that the performance benefits of MTA are architecture-agnostic, effectively enhancing the semantic correctness and helpfulness of diverse student models.

\section{Analysis}
\label{sec:analysis}

To assess the contribution of individual components in MTA, we conduct ablation studies using the \textit{GPT-2 1.5B $\rightarrow$ GPT-2 120M} distillation setting. We analyze two key aspects: (1) the effect of each loss component, and (2) the impact of our layer-adaptive multi-granular alignment strategy. Additional results are reported in Appendix~\ref{sec: Appendix_Ablation}.

\subsection{Impact of Loss Components}
Table~\ref{tab:ablation_loss} evaluates the Dynamic Structural Alignment loss ($\mathcal{L}_{\text{DSA}}$) and the Hidden Representation Alignment loss ($\mathcal{L}_{\text{Hid}}$) when added to FDD, DistiLLM, and DistiLLM-2.

\paragraph{Effect of Individual Losses.}
We observe that adding either $\mathcal{L}_{\text{Hid}}$ or $\mathcal{L}_{\text{DSA}}$ individually improves the average performance across all baselines compared to the corresponding baseline. Adding $\mathcal{L}_{\text{Hid}}$ yields moderate gains by directly aligning intermediate features, improving the average score across all frameworks (e.g., +0.34 for DistiLLM). Introducing $\mathcal{L}_{\text{DSA}}$ often provides larger gains, particularly on benchmarks such as S-NI, indicating that preserving relational structure among semantic units offers complementary supervision beyond point-wise feature matching.
% \begin{itemize}
%     \item \textbf{Effect of $\mathcal{L}_{\text{Hid}}$:} By explicitly aligning feature values via a weighted projection, $\mathcal{L}_{\text{Hid}}$ improves the average ROUGE-L score across all methods (e.g., +0.34 for DistiLLM). This confirms that narrowing the representational gap in the latent space provides a clearer supervision signal than logits alone.
%     \item \textbf{Effect of $\mathcal{L}_{\text{DSA}}$:} The structural alignment consistently outperforms simple feature matching ($\mathcal{L}_{\text{DSA}} > \mathcal{L}_{\text{Hid}}$ in most metrics). For instance, in DistiLLM-2, $\mathcal{L}_{\text{DSA}}$ boosts the S-NI score to 27.06, comparable to the baseline, while significantly improving Dolly and Vicuna. This suggests that preserving the \textit{relational geometry} of the teacher's trajectory is more critical for generalization than merely matching point-wise values.
% \end{itemize}

\paragraph{Synergy of the Full Method.}
Using both losses together achieves the strongest results across all baselines. For example, under DistiLLM, the full MTA configuration improves the average score from 20.21 to 21.45. These results suggest that structural alignment and feature-level alignment capture complementary aspects of the distillation process and are most effective when applied jointly.

\begin{table}[h!]
\centering
\footnotesize
\renewcommand{\arraystretch}{1.15}
\setlength{\tabcolsep}{4.5pt}

\begin{tabular}{l|cccccc}
\toprule
\textbf{Methods} & \textbf{Dolly} & \textbf{SelfInst} & \textbf{Vicuna} & \textbf{S-NI} & \textbf{Avg.}\\
\specialrule{1.0pt}{1.0pt}{1.0pt}

% ================= GPT-2 1.5B -> GPT-2 120M =================
\multicolumn{6}{c}{\textit{GPT-2 1.5B} $\rightarrow$ \textit{GPT-2 120M}}\\
\midrule

FDD            & 25.47 & 12.45 & 16.44 & 23.54 & 19.48 \\
\quad \textit{w/} $\mathcal{L}_{\text{Hid}}$            & 25.39 & 12.42 & 16.9 & 24.18 & 19.72
 \\
\quad \textit{w/} $\mathcal{L}_{\text{DSA}}$            & 25.44	& 12.47 &	17.08 &	24.49 &	19.87
 \\
\quad \textit{w/ Full}      & \underline{25.64} & \underline{13.6} & \underline{17.00} & \underline{25.75} & \underline{20.50} \\
\midrule

DistiLLM       & 25.65 & 13.39 & 16.50 & 25.28 & 20.21 \\
\quad \textit{w/} $\mathcal{L}_{\text{Hid}}$            & \underline{25.89}	& 13.68 &	\underline{16.86} &	25.77 &	20.55
 \\
\quad \textit{w/} $\mathcal{L}_{\text{DSA}}$            & 25.77 & \underline{14.24} &	16.27 &	27.40 & 20.92
 \\
\quad \textit{w/ Full}   & 25.77 & 14.19 & 
                        16.67 & \underline{29.18} & 
                        \underline{21.45} \\
\midrule

DistiLLM\text{-}2      & 22.44 & 12.52 & 12.30 & 27.10 & 18.59 \\
\quad \textit{w/} $\mathcal{L}_{\text{Hid}}$            & \underline{25.26}	& 13.13 &	13.53 &	25.79 &	19.43

 \\
\quad \textit{w/} $\mathcal{L}_{\text{DSA}}$            & 25.08	& 13.28 &	14.09 &	\underline{27.06} & 19.88
 \\
\quad \textit{w/ Full}  & 24.76 & \underline{14.16} & 
                        \underline{14.28} & 26.54 & 
                        \underline{19.94} \\
\bottomrule
\end{tabular}
\caption{Performance of loss combinations on instruction-following benchmarks.}
\label{tab:ablation_loss}
\end{table}

\subsection{Impact of Hierarchical Granularity}
Table~\ref{tab:ablation_level} investigates the validity of our core hypothesis: that distillation should be layer-adaptive, aligning \textit{Word spans} at lower layers and \textit{Phrase spans} at higher layers. We compare our adaptive approach (\textit{Full-level}) against static strategies that enforce uniform granularity (Word-only or Phrase-only) across all layers.

Using a single granularity throughout the network leads to suboptimal performance. A \textbf{word-level-only strategy} is effective for tasks dominated by surface patterns, but it consistently underperforms on reasoning-intensive benchmarks (e.g., Vicuna). For instance, the word-level variant of DistiLLM-2 achieves an average score of 19.10, substantially lower than the adaptive approach. This suggests that enforcing fine-grained alignment at higher layers can restrict the student’s ability to form abstract representations. In contrast, a \textbf{phrase-level-only} strategy generally yields stronger results than word-level alignment (e.g., 21.17 vs.\ 20.80 for DistiLLM), supporting the view that semantic constituents carry richer information. Nevertheless, it still underperforms relative to the full adaptive method, likely because ignoring word-level grounding at lower layers weakens the model’s lexical foundation.

\paragraph{Superiority of Trajectory Alignment.}
The \textit{Full-level} (Layer-Adaptive) strategy consistently achieves the highest scores across all baselines (e.g., 20.50 for FDD and 21.45 for DistiLLM). Notably, on the S-NI benchmark, the adaptive approach in DistiLLM reaches 29.18, outperforming the Phrase-only variant by nearly 2 points. This empirically validates our motivation: aligning the student's representational trajectory to match the teacher's functional hierarchy - moving from lexical grounding to semantic composition - is crucial for distilling capable and robust language models. Furthermore, to assess the contribution of span weights, we conduct additional ablation experiments, the results of which are reported in Appendix~\ref{ablation_weight}.

\begin{table}[h!]
\centering
\footnotesize
\renewcommand{\arraystretch}{1.15}
\setlength{\tabcolsep}{2.5pt}

\begin{tabular}{l|cccccc}
\toprule
\textbf{Methods} & \textbf{Dolly} & \textbf{SelfInst} & \textbf{Vicuna} & \textbf{S-NI} & \textbf{Avg.}\\
\specialrule{1.0pt}{1.0pt}{1.0pt}

% ================= GPT-2 1.5B -> GPT-2 120M =================
\multicolumn{6}{c}{\textit{GPT-2 1.5B} $\rightarrow$ \textit{GPT-2 120M}}\\
\midrule

FDD            & 25.47 & 12.45 & 16.44 & 23.54 & 19.48 \\
\quad \textit{w/ Word-level}             & 25.01 & 12.82 & 17.53 & 24.99 &	20.09
 \\
\quad \textit{w/ Phrase-level}          & 25.29	& 12.87 & \underline{17.36} & 24.55 & 20.02
 \\
\quad \textit{w/ Full-level}      & \underline{25.64} & \underline{13.60} & 17.00 & \underline{25.75} & \underline{20.50} \\
\midrule

DistiLLM       & 25.65 & 13.39 & 16.50 & 25.28 & 20.21 \\
\quad \textit{w/ Word-level}             & 25.82 & 13.54 & 16.67 & 27.16 &	20.80
 \\
\quad \textit{w/ Phrase-level}           & \underline{25.96} & \underline{14.25} & \underline{17.03} & 27.42 &	21.17
 \\
\quad \textit{w/ Full-level}    & 25.77 & 14.19 & 16.67 & \underline{29.18} & \underline{21.45} \\
\midrule

DistiLLM\text{-}2      & 22.44 & 12.52 & 12.30 & 27.10 & 18.59 \\
\quad \textit{w/ Word-level}             & \underline{25.05} & 12.87 & 14.06 & 24.41 &	19.10
 \\
\quad \textit{w/ Phrase-level}            & 24.70 & 13.40 & 14.14 & 25.35 &	19.40
 \\
\quad \textit{w/ Full-level}   & 24.76 & \underline{14.16} & 
                        \underline{14.28} & \underline{26.54} & 
                        \underline{19.94} \\
\bottomrule
\end{tabular}
\caption{Ablations on hierarchical level  mechanisms.}
\label{tab:ablation_level}
\end{table}

\begin{figure}[t]
\centering
\includegraphics[width=\linewidth]{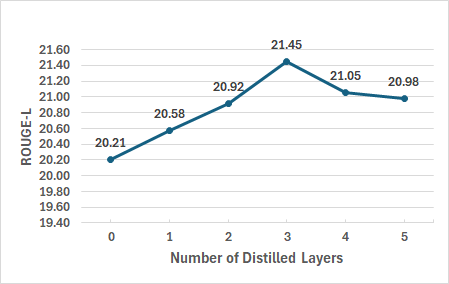}
\captionof{figure}{Effect of the number of distilled intermediate layers. Averaged ROUGE-L scores using DistiLLM + our approach.}
\label{fig:n_distill_layers}
\end{figure}

% \begin{table}[h!]
% \centering
% \footnotesize
% \renewcommand{\arraystretch}{1.15}
% \setlength{\tabcolsep}{3pt}

% \begin{tabular}{l|cccccc}
% \toprule
% \textbf{Word-Phrases} & \textbf{Dolly} & \textbf{Vicuna} & \textbf{SelfInst} & \textbf{S-NI} & \textbf{Avg.}\\
% \specialrule{1.0pt}{1.0pt}{1.0pt}

% % ================= GPT-2 1.5B -> GPT-2 120M =================
% \multicolumn{6}{c}{\textit{GPT-2 1.5B} $\rightarrow$ \textit{GPT-2 120M}}\\
% \midrule

% 0-3            & 25.96 & 14.25 & 17.03 & 27.42 & 21.17 \\
% 1-3            & 25.77 & 14.19 & 16.67 & 29.18 & 21.45 \\
% 2-3            & 25.89 & 14.02 & 16.38 & 27.30 & 20.90 \\
% 3-0            & 25.82 & 13.54 & 16.67 & 27.16 & 20.80 \\
% \bottomrule
% \end{tabular}
% \caption{rate word-phrases with DistiLLM + our method.}
% \label{tab:rate_w_p}
% \end{table}

\begin{table}[h!]
\centering
\footnotesize % Dùng small cho gọn
\renewcommand{\arraystretch}{1.2}
\setlength{\tabcolsep}{2pt}

\begin{tabular}{l|c|cccc|c}
\toprule
\textbf{Config} & \textbf{W:P} & \textbf{Dolly} & \textbf{SelfInst} & \textbf{Vicuna} & \textbf{S-NI} & \textbf{Avg.} \\
\midrule
% Baseline rows
All Phrase & 0 : 3 & 25.96 & 14.25 & 17.03 & 27.42 & 21.17 \\
All Word & 3 : 0 & 25.82 & 13.54 & 16.67 & 27.16 & 20.80 \\
Hybrid A & 2 : 1 & 25.89 & 14.02 & 16.38 & 27.30 & 20.90 \\
\rowcolor{gray!10} \textbf{MTA (Ours)} & \textbf{1 : 2} & 25.77 & 14.19 & 16.67 & 29.18 & \textbf{21.45} \\
\bottomrule
\end{tabular}
\caption{Impact of granularity allocation under the DistiLLM framework with our proposed method. We vary the ratio of Word-level (W) to Phrase-level (P) spans and report performance across benchmarks.}
\label{tab:rate_w_p}
\end{table}
\subsection{Sensitivity Analysis on Layer Selection}
\label{sec:ablation_layer_analysis}

This section analyzes the sensitivity of the distillation process to the number of aligned layers ($M$) and their specific granularity allocation.

\paragraph{Effect of the number of distilled intermediate
layers.} Figure~\ref{fig:n_distill_layers} illustrates the impact of increasing the layer budget $M$ on the student's performance (Avg. ROUGE-L). We observe an inverted U-shaped trend. Performance improves significantly as we increase from $M=0$ (baseline) to $M=3$, reaching a peak score of 21.45. This confirms that sparse intermediate supervision provides necessary guidance for the student's trajectory. However, further increasing $M$ to 4 or 5 leads to diminishing returns and a slight performance drop. This decline supports the hypothesis regarding \textit{inter-layer redundancy}. Thus, we select $M=3$ as the optimal budget for the GPT-2 pair; deeper models use proportionally larger M following the same strided rule (see Table \ref{tab:layer_distill}).

\paragraph{Word vs. Phrase Allocation.}
Fixing the budget at $M=3$, we further analyze how to allocate these layers between Word-level and Phrase-level alignment (Table~\ref{tab:rate_w_p}). Purely structural alignment (\textit{All Phrase}) outperforms purely lexical alignment (\textit{All Word}), highlighting the importance of semantic compositionality. However, the best performance is achieved with a \textbf{Hybrid} configuration (1 Word : 2 Phrase). This empirically validates our hypothesis that the lowest selected layer should be dedicated to lexical grounding, while the upper layers specialize in semantic abstraction. 
\paragraph{Robustness of Layer Schedule.}
To further validate that MTA's improvements are not sensitive to a specific hand-picked layer configuration, we conduct a robustness study on the Qwen1.5 pair (Qwen1.5-1.8B $\rightarrow$ Qwen1.5-0.5B) under multiple word/phrase layer schedules. As described in Section~\ref{layer_mapping}, supervision points are selected via a deterministic strided top-down rule and mapped to teacher layers by proportional scaling, without any per-run search. Table~\ref{tab:layer_robustness} reports results across five alternative schedules, all derived from the same deterministic rule with different stride or budget choices.

Across all configurations, MTA consistently improves over the DistiLLM baseline (Avg. ROUGE-L: 19.80), with gains ranging from +0.75 to +1.21 points. Performance variation across schedules is small (typically $<$0.5 points), confirming that MTA does not rely on a single carefully tuned configuration. In practice, the default strided top-down schedule (Section~\ref{layer_mapping}) is sufficient to reliably improve distillation quality without extensive hyperparameter search. Crucially, no per-run search over word/phrase assignments is needed: our empirical finding that assigning the \textit{lowest} selected layer to word-level 
    and all remaining layers to phrase-level is a robust default 
    that performs competitively across all tested configurations.

\begin{table}[ht]
\centering
\footnotesize
\setlength{\tabcolsep}{1.2pt}
\renewcommand{\arraystretch}{1.3}
\begin{tabular}{llccccc}
\toprule
\textbf{Word-level} & \textbf{Phrase-level} & \textbf{Dolly} & \textbf{SelfInst} & \textbf{Vicuna} & \textbf{S-NI} & \textbf{Avg.} \\
\midrule
\multicolumn{2}{l}{\textit{DistiLLM (baseline)}} & 25.16 & 12.90 & 15.86 & 25.28 & 19.80 \\
\midrule
14            & 16--24  & 25.61 & 13.08 & 16.04 & \textbf{29.32} & \textbf{21.01} \\
14, 16, 18    & 20, 22, 24          & 25.30 & 13.02 & 16.05 & 29.10 & 20.87 \\
16, 18, 20    & 22, 24              & 25.51 & \textbf{13.41} & \textbf{16.27} & 27.02 & 20.55 \\
19            & 20--24  & \textbf{26.40} & \underline{14.63} & 16.08 & 25.54 & 20.66 \\
12, 14, 16    & 18--24    & 25.11 & 13.11 & \underline{16.50} & \underline{29.17} & \underline{20.97} \\
\bottomrule
\end{tabular}
\caption{
    Sensitivity of MTA to different layer schedule configurations on Qwen1.5-1.8B $\rightarrow$ Qwen1.5-0.5B. Here we only use even layers.
}
\label{tab:layer_robustness}
\end{table}
\subsection{Computational Efficiency Analysis}
Table~\ref{tab:efficiency_main} reports the per-step wall-clock time and GPU memory consumption of MTA compared to the baselines. All experiments are conducted on a single NVIDIA A100 (40\,GB) GPU with a batch size of 16. The additional training overhead in MTA stems from the syntactic span extraction stage (via spaCy) and the span-level geometry matching in $\mathcal{L}_{\text{DSA}}$. 

In our initial implementation, span extraction contributed 0.42\,s/step, resulting in a 2.54$\times$ slowdown over DistiLLM. After optimizing the extraction pipeline via larger spaCy batch sizes, this overhead is reduced to 0.27\,s/step, bringing the overall slowdown down to 1.85$\times$ with no increase in GPU memory usage. Importantly, \textbf{MTA adds overhead only during training}; inference cost remains completely unchanged since span extraction is not used at test time.

To verify that the performance gains are not merely a consequence of increased training budget, we conduct a time-matched comparison where the baselines are trained for $3\times$ more epochs, granting them a wall-clock budget equal to or exceeding that of MTA. As shown in Table~\ref{tab:time_matched}, even with substantially more training time, the baselines do not close the performance gap, confirming that MTA's improvements stem from the proposed trajectory-alignment objective rather than extra compute.

\begin{table}[ht]
\small
\renewcommand{\arraystretch}{1.0}
\setlength{\tabcolsep}{1.0pt}
\begin{tabular}{lcccc}
\toprule
\textbf{Method} & \textbf{Span Extr.} & \textbf{Time} & \textbf{avg\_alloc} & \textbf{peak\_alloc} \\
\midrule
DistiLLM                          & 0.00 & 0.26 & 6.53 & 16.91 \\
% DistiLLM + MTA                    & 0.42 & 0.66 & 6.54 & 17.94 \\
DistiLLM + MTA       & 0.27 & 0.48 & 6.54 & 17.94 \\
\midrule
FDD                               & 0.00 & 0.49 & 6.67 & 23.04 \\
% FDD + MTA                         & 0.51 & 0.88 & 6.70 & 24.05 \\
FDD + MTA               & 0.36 & 0.69 & 6.70 & 24.05 \\
\bottomrule
\end{tabular}
\caption{
    Computation time and GPU memory consumption. 
    MTA introduces overhead only during training; inference cost is unchanged.
}
\label{tab:efficiency_main}
\end{table}

\begin{table}[ht]
\centering
\footnotesize
\renewcommand{\arraystretch}{1.1}
\setlength{\tabcolsep}{2.0pt}
\begin{tabular}{lccccc}
\toprule
\textbf{Training Setup} & \textbf{Dolly} & \textbf{SelfInst} & \textbf{Vicuna} & \textbf{S-NI} & \textbf{Avg.} \\
\midrule
\multicolumn{6}{c}{DistiLLM variants} \\
\midrule
Train w/ 5 ep           & 25.65 & 13.39 & 16.50 & 25.28 & 20.21 \\
Train w/ 10 ep          & 26.64 & 13.11 & 17.49 & 23.83 & 20.27 \\
Train w/ 15 ep          & 26.61 & 13.35 & 16.96 & 24.13 & 20.26 \\
 w/ MTA (5 ep)     & \textbf{25.77} & \textbf{14.19} & \textbf{16.67} & \textbf{29.18} & \textbf{21.45} \\
\midrule
\multicolumn{6}{c}{DistiLLM-2 variants} \\
\midrule
Train w/ 5 ep         & 22.44 & 12.52 & 12.30 & 27.10 & 18.59 \\
Train w/ 10 ep        & 22.96 & 12.51 & 12.88 & 27.91 & 19.07 \\
Train w/ 15 ep        & 22.97 & 12.46 & 13.09 & 27.68 & 19.05 \\
w/ MTA (5 ep)   & \textbf{24.76} & \textbf{14.16} & \textbf{14.28} & \textbf{26.54} & \textbf{19.94} \\
\bottomrule
\end{tabular}
\caption{
    Time-matched comparison on GPT-2 1.5B $\rightarrow$ GPT-2 120M.
    Baselines trained for up to 15 epochs (3$\times$ the wall-clock budget of MTA at 5 epochs) 
    do not close the performance gap.
}
\label{tab:time_matched}
\end{table}
\section{Conclusion}
We introduce \textbf{Multi-Granular Trajectory Alignment (MTA)}, a layer-adaptive distillation framework that models the hierarchical representational trajectory of large language models.
MTA aligns word-level representations in lower layers and phrase-level representations in higher layers, matching the depth-dependent semantic roles of Transformer layers.
By combining structural and hidden-state alignment objectives, MTA enables students to preserve both local features and global relational geometry, yielding consistent gains across strong baselines.

\section{Limitations}
While MTA improves distillation quality, it introduces additional computational cost due to the use of external sparse structures for layer-wise alignment. Although this overhead remains manageable in our current experiments, an important direction for future work is to design more lightweight yet still layer-adaptive alignment mechanisms that reduce computational cost while preserving performance. In addition, our experiments are conducted under fixed computational budgets and benchmark settings. We view these constraints as opportunities for further development, particularly in exploring more efficient layer-wise alignment designs and adaptive mapping strategies that can enhance the scalability and robustness of MTA.

\section*{Acknowledgments}
Trung Le was supported by the Air Force Office of Scientific Research under award number FA2386-25-1-4023 and the ARC Discovery Project grant DP250100262.

%MTA introduces 1 cái chi phí tính toán vì hiện tại chúng tôi vẫn đang sử dụng sparse ngoài để tlính. Fureture work về việc tìm cách catch cái layerwise mà nó đỡ chậm hơn.
% MTA introduces additional computational overhead compared to standard point-wise distillation objectives. Specifically, the extraction of multi-granular spans and the computation of pairwise relational distances among spans increase both memory usage and training time, particularly when aligning higher layers with a large number of phrase spans. This cost grows with the number of spans per layer, which may limit scalability for very long input sequences. In practice, this trade-off can be mitigated by restricting alignment to a small set of key layers or pruning low-salience spans. Moreover, MTA relies on externally extracted spans, which may introduce sensitivity to the quality of the parsing or span selection method, especially for domains or languages with limited syntactic resources. We leave the exploration of more efficient span selection and parser-free alternatives as future work.

\bibliography{ref}
\bibliographystyle{acl_natbib}

\newpage
\appendix
\newpage
\appendix
\section{Baseline models}
\label{sec: appendix_baseline}
\subsection{DistiLLM}
To address the optimization instability and high computational costs associated with standard KD and on-policy data generation, \citet{ko2024distillm} proposed DistiLLM. This framework introduces two primary innovations:

\noindent \textbf{Skew (Reverse) Kullback-Leibler Divergence:}
Standard KLD often leads to optimization instability due to gradient explosion when the student model assigns low probability to the teacher's support. To mitigate this, DistiLLM utilizes \textbf{Skew KLD}, which interpolates the target distribution with the student's distribution. Formally, the $\alpha$-Skew KLD is defined as:
\begin{equation}
    D_{SKL}^{(\alpha)}(p, q_\theta) = D_{KL}(p \parallel \alpha p + (1-\alpha)q_\theta)
\end{equation}
Similarly, to leverage the mode-seeking behavior of Reverse KLD while maintaining gradient stability, they proposed \textbf{Skew Reverse KLD (SRKL)}:
\begin{equation}
    D_{SRKL}^{(\alpha)}(p, q_\theta) = D_{KL}(q_\theta \parallel (1-\alpha)p + \alpha q_\theta)
\end{equation}
where $p$ is the teacher, $q_\theta$ is the student, and $\alpha \in [0, 1]$ controls the mixing ratio. These modifications prevent the denominator in the gradient term from vanishing, thereby bounding the gradient norm and ensuring smoother convergence.

\noindent \textbf{Adaptive Off-Policy Generation:}
While utilizing Student-Generated Outputs (SGOs) addresses training-inference mismatch, generating sequences at every iteration (on-policy) is computationally expensive. DistiLLM proposes an efficient off-policy strategy using a replay buffer $\mathcal{D}_R$ to store and reuse SGOs. Furthermore, it employs an \textit{adaptive SGO scheduler} that dynamically adjusts the probability $\phi$ of sampling SGOs based on the student's validation loss trends. This mechanism ensures SGOs are introduced only when necessary to correct distribution shifts, balancing training efficiency with the mitigation of noisy feedback.
\subsection{DistiLLM-2}
\label{subsec: appendix_A2}
Building upon the success of Skew KLD and adaptive data strategies, \citet{ko2025distillm2} introduced DistiLLM-2, a contrastive framework designed to align teacher and student models more effectively across varied data types. The method comprises three key technical advancements:

\noindent \textbf{Contrastive Approach for LLM Distillation (CALD):}
Motivated by the observation that Forward KL and Reverse KL exhibit asymmetric behaviors—"pulling up" probabilities on teacher-generated outputs (TGOs) and "pushing down" on student-generated outputs (SGOs), respectively—DistiLLM-2 proposes a dual-loss objective. It applies Skew KL (SKL) to TGOs ($y_t$) to encourage matching high-probability regions, and Skew Reverse KL (SRKL) to SGOs ($y_s$) to suppress low-quality generations. The combined objective is defined as:
\begin{multline}
\small
  \mathcal{L}_{CALD} = \frac{1}{2|\mathcal{D}|} \sum_{(x, y_t, y_s) \sim \mathcal{D}} \Big[ (1-\beta) D_{SKL}^{(\alpha_t)}(x, y_t) \\
  + \beta D_{SRKL}^{(\alpha_s)}(x, y_s) \Big]
\end{multline}
where $\beta$ balances the two terms. This formulation is mathematically connected to preference optimization methods like DPO but tailored for distillation to avoid reward hacking.

\noindent \textbf{Curriculum-based Adaptive Learning:}
To further refine the objective, DistiLLM-2 introduces dynamic updates for the skew coefficient $\alpha$. Recognizing that the optimal mixing ratio depends on the sample difficulty (the gap between teacher and student distributions), the method employs a curriculum-based update rule derived from a first-order Taylor approximation:
\begin{equation}
    \alpha \approx 1 - (1 - \alpha_0) \frac{m}{p(y|x) - q_\theta(y|x)}
\end{equation}
Additionally, the coefficient $\beta$ for the SRKL term is gradually increased during training, shifting focus from imitating the teacher to correcting student errors as the model capability improves.

\noindent \textbf{Optimal Data Curation:}
Instead of purely on-policy generation, DistiLLM-2 adopts a "batched" data curation strategy where SGOs are collected ahead of each training epoch. This approach maintains the benefits of on-policy feedback while significantly improving computational efficiency and compatibility with high-throughput inference engines like vLLM.

\section{Extended Related Work}
\label{sec:related_work}

\paragraph{Knowledge distillation for language models.}
Knowledge distillation (KD) trains a compact student to imitate a larger teacher, most commonly by matching the teacher’s output distribution via KL-based objectives \citep{hinton2015distillingknowledgeneuralnetwork,kim2016sequencelevelknowledgedistillation}.
In the context of large language models (LLMs), logit-level KD remains a strong baseline, but can suffer from optimization instability and train-test mismatch when trained purely under teacher forcing \citep{agarwal2024onpolicy}.
Recent frameworks improve practicality and stability through better divergence formulations and data strategies \citep{ko2024distillm,ko2025distillm2,agarwal2024onpolicy,anshumann2025sparse}.
These methods primarily focus on aligning behavior at the output level, which is effective but may under-constrain the internal knowledge encoded in intermediate representations.

\paragraph{Intermediate representation alignment and trajectory-based distillation.}
A complementary line of work distills internal signals - hidden states, attention maps, and other intermediate features - to transfer information beyond output logits.
Representative Transformer distillation methods align intermediate hidden states and/or attentions \citep{jiao2020tinybert,sun2019patientknowledgedistillationbert,sun2020mobilebertcompacttaskagnosticbert}, while others emphasize distilling attention relations to build smaller but competitive students \citep{wang2020minilmdeepselfattentiondistillation}.
Recent work also revisits intermediate supervision for LLM distillation under efficiency constraints, often choosing a sparse subset of layers \citep{gong2025beyond,song2025demystifyingrolesllmlayers, an2026mol}.

Beyond matching isolated layers, trajectory-style methods aim to match how representations change across depth.
For example, Feature Dynamics Distillation (FDD) explicitly treats intermediate features as a discretized depth-wise path and aligns not only the feature trajectory but also its first-order dynamics \citep{gong2025beyond}.
Trajectory ideas also appear at the training-process level: BERT-of-Theseus progressively replaces teacher modules with compact substitutes during training \citep{xu2020bertoftheseus}.
Separately, some work studies ``trajectory'' over generated tokens by distilling step-by-step rationales \citep{wang2023scott,li2023scotd}.

\paragraph{Hierarchy in Transformer representations and linguistically motivated compression.}
A substantial body of work shows that Transformer layers form a functional hierarchy, where lower layers encode surface/lexical information and higher layers capture increasingly abstract and semantic features \citep{tenney2019bert,rogers2020primer,clark2019what}.
Linguistics similarly characterizes language as hierarchical composition from lexical units into phrases and higher-level structures \citep{chomsky1965aspects,crain1987structure,socher2011parsing,socher2013recursive}.
Recent analyses further suggest that modern LLMs exhibit depth-specialized behaviors such as memory-like retrieval in lower layers and more abstract reasoning in higher layers \citep{wang2025think,yang2025internal}.
In addition, representational \emph{geometry} has been shown to encode separable syntactic/semantic structure and fine-grained organization of linguistic features in BERT \citep{coenen2019visualizingmeasuringgeometrybert}, further motivating distillation objectives that preserve relational structure rather than only per-token values.
These findings motivate distillation objectives that respect hierarchical structure.
In our work, we instantiate this idea as \emph{Multi-Granular Trajectory Alignment (MTA)}, which performs depth-adaptive alignment: word-level units for lower layers and phrase-level spans (e.g., NPs/VPs) for higher layers, thereby constraining not only feature values but also the evolving relational geometry across depth.

MTA complements both logit-based and intermediate-representation KD by adding (i) \emph{Dynamic Structural Alignment} that matches within-layer relational structure among semantic units, and (ii) \emph{Hidden Representation Alignment} for selected layers, with salience-aware weighting.
Unlike depth-uniform feature matching, our layer-adaptive design explicitly tracks the hierarchical representational trajectory, aligning lexical grounding early and compositional semantics later, and can be plugged into recent LLM distillation pipelines \citep{ko2024distillm,ko2025distillm2,gong2025beyond}.

\section{Token Importance Computation}
\label{sec:appendix_token_weights}

To compute the token-level importance weights $w \in \mathbb{R}^N$, we employ a standardized pairwise self-attention mechanism that captures the global contextual relevance of each token while explicitly excluding self-loops. Let $\mathbf{H}_l = [\mathbf{h}_{1, l}, \dots, \mathbf{h}_{N, l}] \in \mathbb{R}^{N \times d}$ denote the sequence of hidden states at layer $l$-th, where $N$ is the sequence length and $d$ is the feature dimension.

First, to stabilize the dot-product scores, we standardize the hidden states to obtain $\hat{\mathbf{H}}$, where each vector is scaled by its standard deviation:
\begin{equation}
    \hat{H}_{t,l} = \frac{H_{t,l}}{\sigma(H_{t,l})},
\end{equation}
where $\sigma(\cdot)$ computes the standard deviation along the feature dimension.

Next, we calculate the pairwise attention scores $\mathbf{S}_l \in \mathbb{R}^{N \times N}$. We incorporate a masking term $\mathbf{M}$ to enforce two constraints: (1) tokens cannot attend to padding elements, and (2) tokens cannot attend to themselves (diagonal masking), forcing the model to rely solely on context:
\begin{equation}
    S_{s\to t,l} = \frac{\hat{H}_{s,l} \hat{H}_{t,l}^\top}{\sqrt{d}} + M_{s,t},
\end{equation}
where the mask $M_{s,t}$ enforces both padding and diagonal constraints:
\begin{equation}
    M_{s,t} =
    \begin{cases}
        -\infty, & s = t \text{ or } t \in \mathcal{P}, \\
        0, & \text{otherwise},
    \end{cases}
\end{equation}
where $\mathcal{P}$ represents the set of padding indices.

The attention weights are obtained via the softmax function. Finally, the scalar importance weight $w_t$ for the $t$-th token is computed as the mean attention it receives from all other tokens in the sequence (column-wise average):
\begin{equation}
    \alpha_{s\to t,l} =
    \frac{\exp(S_{s\to t,l})}
    {\sum_{u=1}^{N} \exp(S_{s\to u,l})}.
\end{equation}
\begin{equation}
    w_{t,l} = \frac{1}{N} \sum_{s=1}^{N} \alpha_{s\to t,l}.
\end{equation}
The resulting vector $\mathbf{w}_l = [w_{1, l}, \dots, w_{N, l}]$ represents the computed token weights at layer $l$-th.

\section{Experimental Details}
\label{sec: appendix_exp}

\paragraph{Training and Evaluation } For GPT-2 and Qwen1.5 models, we perform full-parameter fine-tuning, while for OPT models we adopt LoRA-based parameter-efficient fine-tuning. Detailed training configurations for each model are summarized in Table~\ref{tab:training-config} and Table~\ref{tab:training-config-2}. 
For evaluation, we sample model outputs using five different random seeds to account for stochasticity in generation. The final performance is reported as the average ROUGE-L score \cite{lin2004rouge} between the generated responses and the human-annotated references.

\begin{table}[htbp]
\centering
\setlength{\tabcolsep}{1pt}
\small
\begin{tabular}{l|ccc}
\toprule
\textbf{Settings} & \textbf{GPT2-1.5B} & \textbf{Qwen1.5-1.8B} & \textbf{OPT-6.7B} \\
\cmidrule(lr){2-2} \cmidrule(lr){3-3} \cmidrule(lr){4-4}
& GPT2-120M & Qwen1.5-0.5B & OPT-1.3B \\
\midrule
Epoch          & 5 & 5 & 5 \\
LR             & $1\times10^{-4}$ & $1\times10^{-4}$ & $5\times10^{-4}$ \\
Projector LR   & $5\times10^{-4}$ & $5\times10^{-4}$ & $5\times10^{-4}$ \\
Batch Size     & 16 & 16 & 16 \\
LR Scheduler   & Cosine & Cosine & Cosine \\
Fine-Tuning    & Full & Full & LoRA \\
LoRA Rank      & -- & -- & 16 \\
LoRA Alpha     & -- & -- & 64 \\
LoRA Dropout   & -- & -- & 0.1 \\
\bottomrule
\end{tabular}
\caption{Training configurations for FDD and DistiLLM.}
\label{tab:training-config}
\end{table}

\begin{table}[htbp]
\centering
\setlength{\tabcolsep}{1pt}
\small
\begin{tabular}{l|ccc}
\toprule
\textbf{Settings} & \textbf{GPT2-1.5B} & \textbf{Qwen1.5-1.8B} & \textbf{OPT-6.7B} \\
\cmidrule(lr){2-2} \cmidrule(lr){3-3} \cmidrule(lr){4-4}
& GPT2-120M & Qwen1.5-0.5B & OPT-1.3B \\
\midrule
Epoch          & 5 & 5 & 3 \\
LR             & $1\times10^{-4}$ & $5\times10^{-5}$ & $5\times10^{-4}$ \\
Projector LR   & $5\times10^{-4}$ & $5\times10^{-4}$ & $5\times10^{-4}$ \\
Batch Size     & 8 & 8 & 8 \\
LR Scheduler   & Cosine & Cosine & Cosine \\
Fine-Tuning    & Full & Full & LoRA \\
LoRA Rank      & -- & -- & 16 \\
LoRA Alpha     & -- & -- & 128 \\
LoRA Dropout   & -- & -- & 0.1 \\
\bottomrule
\end{tabular}
\caption{Training configurations for DistiLLM-2.}
\label{tab:training-config-2}
\end{table}

\paragraph{Hyperparameter}
The hyperparameters, $\lambda_{\text{DSA}}$ and $\lambda_{\text{Hid}}$, are reported in Table~\ref{tab:loss_coeff}. All reported values are selected based on preliminary validation experiments.

\begin{table}[hb]
\centering
\setlength{\tabcolsep}{1.0pt}
\small
\begin{tabular}{lcccccc}
\toprule
\textbf{Method} 
& \multicolumn{2}{c}{\textbf{GPT2-120M}} 
& \multicolumn{2}{c}{\textbf{Qwen1.5-0.5B}} 
& \multicolumn{2}{c}{\textbf{OPT-1.3B}} \\
\cmidrule(lr){2-3} \cmidrule(lr){4-5} \cmidrule(lr){6-7}
& $\lambda_{\text{DSA}}$ & $\lambda_{\text{Hid}}$ & $\lambda_{\text{DSA}}$ & $\lambda_{\text{Hid}}$ & $\lambda_{\text{DSA}}$ & $\lambda_{\text{Hid}}$ \\
\midrule
FDD + Ours         & 2 & 0.2 & 2 & 0.2 & 2 & 0.2 \\
DistiLLM + Ours  & 2 & 0.2 & 2 & 0.2 & 3 & 0.3 \\
DistiLLM-2 + Ours & 2 & 0.2 & 2 & 0.2 & 3 & 0.3 \\
\bottomrule
\end{tabular}
\caption{Loss weighting coefficients used for different methods and model sizes.}
\label{tab:loss_coeff}
\end{table}

\paragraph{Layer Mapping Configuration.}
\label{layer_mapping}
Table~\ref{tab:layer_distill} summarizes the selected intermediate layers used for distillation at the word and phrase levels.
For models with different depths, we align student layers to teacher layers using a fixed integer mapping.
Motivated by the high inter-layer redundancy observed in Transformers \citep{song2025demystifyingrolesllmlayers, gong2025beyond}, where adjacent layers often encode similar features, we avoid computationally expensive full-depth alignment. Instead, we select a concise subset of \textit{key layers} $\mathcal{L}_{\text{key}}$ to efficiently capture how representations evolve across layers.
To address the depth mismatch between the Teacher ($N_{\mathcal{T}}$) and Student ($N_{\mathcal{S}}$), we map each selected Student layer $l_{\mathcal{S}}$ to its corresponding Teacher layer $l_{\mathcal{T}}$ via proportional scaling: $l_{\mathcal{T}} = \left\lfloor l_{\mathcal{S}} \times \frac{N_{\mathcal{T}}}{N_{\mathcal{S}}} \right\rfloor$.
We define $\mathcal{L}_{\text{key}}$ using a strided top-down approach to prioritize critical high-level semantic transitions near the output. Given a stride $k$ and a layer budget $M$, the selected Student indices are:
    $\mathcal{I}_{\mathcal{S}} = \{ N_{\mathcal{S}}, N_{\mathcal{S}} - k, \dots \}\; \text{ with} \; |\mathcal{I}_{\mathcal{S}}| = M $. This discretization (typically $k \in \{2, 3\}$) effectively filters out redundant intermediate states while maintaining a coherent evolutionary trajectory. Note that while M = 3 is selected as the optimal budget for the GPT-2 pair (Section \ref{sec:ablation_layer_analysis}), deeper architectures such as Qwen1.5-0.5B and OPT-1.3B use a larger budget (M = 5–6) to provide sufficient supervision points along their longer representational trajectories. The budget M scales naturally with model depth via the strided top-down rule.

\begin{table}[ht]
\centering
\small
\setlength{\tabcolsep}{1.0pt}
\begin{tabular}{lcc}
\toprule
\textbf{Model} & \multicolumn{2}{c}{\textbf{Layer Distillation}} \\
\cmidrule(lr){2-3}
 & \textbf{Word Level} & \textbf{Phrase Level} \\
\midrule
GPT-2 120M   & 6  & 9, 12 \\
Qwen1.5 0.5B & 14 & 16, 18, 20, 22, 24 \\
OPT 1.3B    & 16 & 18, 20, 22, 24 \\
\bottomrule
\end{tabular}
\caption{Selected intermediate layers for word-level and phrase-level distillation across different models.}
\label{tab:layer_distill}
\end{table}

% \paragraph{Granularity Assignment Strategy.}

\section{Additional Ablation Results}
\label{sec: Appendix_Ablation}

% \paragraph{Computational Efficiency and GPU Memory.} All experiments are conducted using GPT-2 XL and GPT-2 120M models on a single NVIDIA A100 GPU with (40\,GB) memory. We use a batch size of 16 for all settings. As shown in Table~\ref{tab:efficiency}, the increased computation time of our method mainly stems from the use of a syntactic parsing tree.
% While this introduces additional overhead, it allows the model to leverage richer structural information, resulting in better overall performance.

% \begin{table}[ht]
% \centering
% % \small
% \begin{tabular}{lccc}
% \toprule
% \textbf{Methods} & \textbf{Time/step} & \textbf{avg\_alloc (GB)} & \textbf{peak\_alloc (GB)} \\
% \midrule
% DistiLLM        & 0.26 & 6.53 & 16.91 \\
% DistiLLM + Ours & 0.66 & 6.54 & 17.94 \\
% FDD             & 0.49 & 6.67 & 23.04 \\
% FDD + Ours     & 0.88 & 6.70 & 24.05 \\
% \bottomrule
% \end{tabular}
% \caption{Computation time and GPU memory consumption of different methods.}
% \label{tab:efficiency}
% \end{table}

\paragraph{Importance of Span Weights.}
\label{ablation_weight}
To validate the necessity of our weighting mechanism, we compare our aggregation strategy against a baseline using uniform mean pooling (denoted as \textit{w/o weight}). As shown in Table~\ref{tab:ablation_weight}, removing the importance weights leads to a consistent performance degradation across both distillation frameworks.

Specifically, for the DistiLLM backbone, incorporating the weighting mechanism boosts the average ROUGE-L score from 20.66 to 21.45. Notably, performance on the \textit{Super-Natural Instructions (S-NI)} benchmark sees a substantial improvement (+2.97 points). This indicates that not all tokens within a span contribute equally to its semantic representation. By leveraging the teacher's attention patterns to suppress non-informative tokens (e.g., stopwords or padding) and highlight salient ones, our method allows the student to capture a more precise, focused, and semantically consistent representational trajectory.
\begin{table}[h!]
\centering
\small
\renewcommand{\arraystretch}{1.1}
\setlength{\tabcolsep}{1pt}
\begin{tabular}{l|ccccc}
\toprule
\textbf{Methods} & \textbf{Dolly} & \textbf{SInst} & \textbf{Vicuna} & \textbf{S-NI} & \textbf{Avg.}\\
\specialrule{1.0pt}{1.0pt}{1.0pt}
\multicolumn{6}{c}{\textit{GPT-2 1.5B} $\rightarrow$ \textit{GPT-2 120M}}\\
\midrule
FDD + Ours \textit{w/o}    & 25.09 & 12.44 & \underline{17.03} & 25.17 & 19.93 \\
FDD + Ours \textit{w/}     & \underline{25.64} & \underline{13.60} & 17.00 & \underline{25.75} & \underline{20.50} \\
\midrule
DistiLLM + Ours \textit{w/o}   & \underline{25.95} & 14.10 & 16.38 & 26.21 & 20.66 \\
DistiLLM + Ours \textit{w/}    & 25.77 & \underline{14.19} & \underline{16.67} & \underline{29.18} & \underline{21.45} \\
\bottomrule
\end{tabular}
\caption{Ablations on span weight mechanisms. \textit{w/} and \textit{w/o} denote with and without importance weighting.}
\label{tab:ablation_weight}
\end{table}

\section{Prompt for evaluation via GPT-4}
\label{appendix:prompt}
\begin{figure}[htbp]
\centering
\begin{tcolorbox}[colback=promptbg, colframe=blue, boxrule=0.5pt, 
                  left=3pt,right=3pt,top=3pt,bottom=3pt,
                  width=1\columnwidth]
\scriptsize 
\raggedright

\textbf{[System]}
\begin{itemize}[leftmargin=*, noitemsep, topsep=0pt]
    \item Please act as an impartial judge and evaluate the quality of the response provided by an AI assistant to a query taken from the test dataset displayed below.
    \item A ground truth answer is provided and should be treated as the correct reference.
    \item Assess whether the assistant’s response is accurate compared to the ground truth and whether the wording and explanation are appropriate and coherent.
    \item Begin your evaluation by providing a short explanation. Be as objective as possible.
    \item After providing your explanation, please rate the response on a scale of 1 to 100 by strictly following this format: ``[[rating]]'', for example: ``Rating: [[90]]''.
\end{itemize}

\medskip
\textbf{[Question]}\\
\{question\}

\medskip
\textbf{[Ground truth answer]}\\
\{ground truth answer\}

\medskip
\textbf{[The Start of Assistant’s Answer]}\\
\{assistant response\}\\
\textbf{[The End of Assistant’s Answer]}

\end{tcolorbox}
\caption{Prompt for GPT-4 evaluation using Ground Truth.}
\label{fig:gpt4_eval_gt}
\vspace{-2mm}
\end{figure}

\cleardoublepage

\end{document}